\documentclass[10pt,twocolumn,letterpaper]{article}

\usepackage{cvpr}      

\usepackage{graphicx}
\usepackage{amsmath}
\usepackage{amssymb}
\usepackage{booktabs}
\usepackage{times}
\usepackage{epsfig}
\usepackage{amssymb}
\usepackage{makecell}
\usepackage{multirow}
\usepackage{algorithm}
\usepackage{caption}
\usepackage{bm}
\usepackage{colortbl}
\usepackage{wrapfig}
\usepackage{pifont}

\usepackage{xcolor}
\usepackage{listings}
\usepackage[british,american]{babel}
\usepackage{enumitem}

\usepackage{textpos}  

\newlength\savedwidth
\newcommand\whline{\noalign{\global\savedwidth\arrayrulewidth\global\arrayrulewidth 0.8pt}\hline\noalign{\global\arrayrulewidth\savedwidth}}

\newcommand{\green}[1]{\textcolor[RGB]{96,177,87}{#1}}
\newcommand{\fn}[1]{\footnotesize{#1}}
\newcommand{\gbf}[1]{\green{\bf{\fn{(#1)}}}}

\definecolor{mygray}{gray}{.92}

\setitemize[1]{itemsep=0pt,partopsep=0pt,parsep=\parskip,topsep=5pt}

\newlength\savewidth

\makeatletter\renewcommand\paragraph{\@startsection{paragraph}{4}{\z@}
  {.5em \@plus1ex \@minus.2ex}{-.5em}{\normalfont\normalsize\bfseries}}\makeatother

\newcommand{\app}{\raise.17ex\hbox{$\scriptstyle\sim$}}

\usepackage{tabulary}
\newcolumntype{x}[1]{>{\centering\arraybackslash}p{#1pt}}
\usepackage{etoolbox}
\makeatletter
\AfterEndEnvironment{algorithm}{\let\@algcomment\relax}
\AtEndEnvironment{algorithm}{\kern2pt\hrule\relax\vskip3pt\@algcomment}
\let\@algcomment\relax
\newcommand\algcomment[1]{\def\@algcomment{\footnotesize#1}}
\renewcommand\fs@ruled{\def\@fs@cfont{\bfseries}\let\@fs@capt\floatc@ruled
  \def\@fs@pre{\hrule height.8pt depth0pt \kern2pt}%
  \def\@fs@post{}%
  \def\@fs@mid{\kern2pt\hrule\kern2pt}%
  \let\@fs@iftopcapt\iftrue}
\makeatother

\usepackage[pagebackref,breaklinks,colorlinks]{hyperref}
 
\makeatletter
\newcommand{\thickhline}{%
    \noalign {\ifnum 0=`}\fi \hrule height 1.2pt
    \futurelet \reserved@a \@xhline
}

\usepackage[capitalize]{cleveref}
\crefname{section}{Sec.}{Secs.}
\Crefname{section}{Section}{Sections}
\Crefname{table}{Table}{Tables}
\crefname{table}{Tab.}{Tabs.}

\def\cvprPaperID{1485} 
\def\confName{CVPR}
\def\confYear{2023}
\newcommand\ourmethod{TT}

\begin{document}

\title{Token Transformer: Can class token help window-based transformer build better long-range interactions?}

\author{Jiawei Mao \quad Yuanqi Chang  \quad Xuesong Yin {\thanks{Corresponding author.}}  \\ 
School of Media and Design, Hangzhou Dianzi University, Hangzhou, China \qquad \\
{\tt\small\{jiaweima0,211330020,yinxs\}@hdu.edu.cn }\\
}

\maketitle

\begin{abstract}
  Compared with the vanilla transformer, the window-based transformer offers a better trade-off between accuracy and efficiency. 
  Although the window-based transformer has made great progress, its long-range modeling capabilities are limited due to the size of the local window and the window connection scheme. 
  To address this problem, we propose a novel Token Transformer (TT). The core mechanism of TT is the addition of a Class (CLS) token for summarizing window information in each local window. 
  We refer to this type of token interaction as CLS Attention. These CLS tokens will interact spatially with the tokens in each window to enable long-range modeling. 
  In order to preserve the hierarchical design of the window-based transformer, we designed Feature Inheritance Module (FIM) in each phase of TT to deliver the local window information from the previous phase to the CLS token in the next phase. 
  In addition, we have designed a Spatial-Channel Feedforward Network (SCFFN) in TT, which can mix CLS tokens and embedded tokens on the spatial domain and channel domain without additional parameters. 
  Extensive experiments have shown that our TT achieves competitive results with low parameters in image classification and downstream tasks. 
\end{abstract}

\section{Introduction}

In recent years, Vision Transformer(ViT) \cite{dosovitskiy2020image} has relied on attention mechanisms for global contextual modeling and its own architecture to achieve success in various vision tasks. 
Compared with Convolutional Neural Network (CNN), various ViT improvement efforts \cite{touvron2021training,zhou2021deepvit,touvron2021going,chen2021crossvit,graham2021levit,wu2021cvt,chu2021twins,wang2021crossformer,tu2022maxvit,li2022sepvit,tang2022quadtree,pan2021scalable,pan2022fast,d2021convit,li2021localvit} have achieved results that can compete with them on multiple datasets. 
Although ViT performs well in computer vision, the secondary computational complexity caused by attention leads to the huge computational overhead of ViT. 
This has hindered the further development of ViT.

\begin{figure}[t]
  \centering
  \includegraphics[width=1\linewidth]{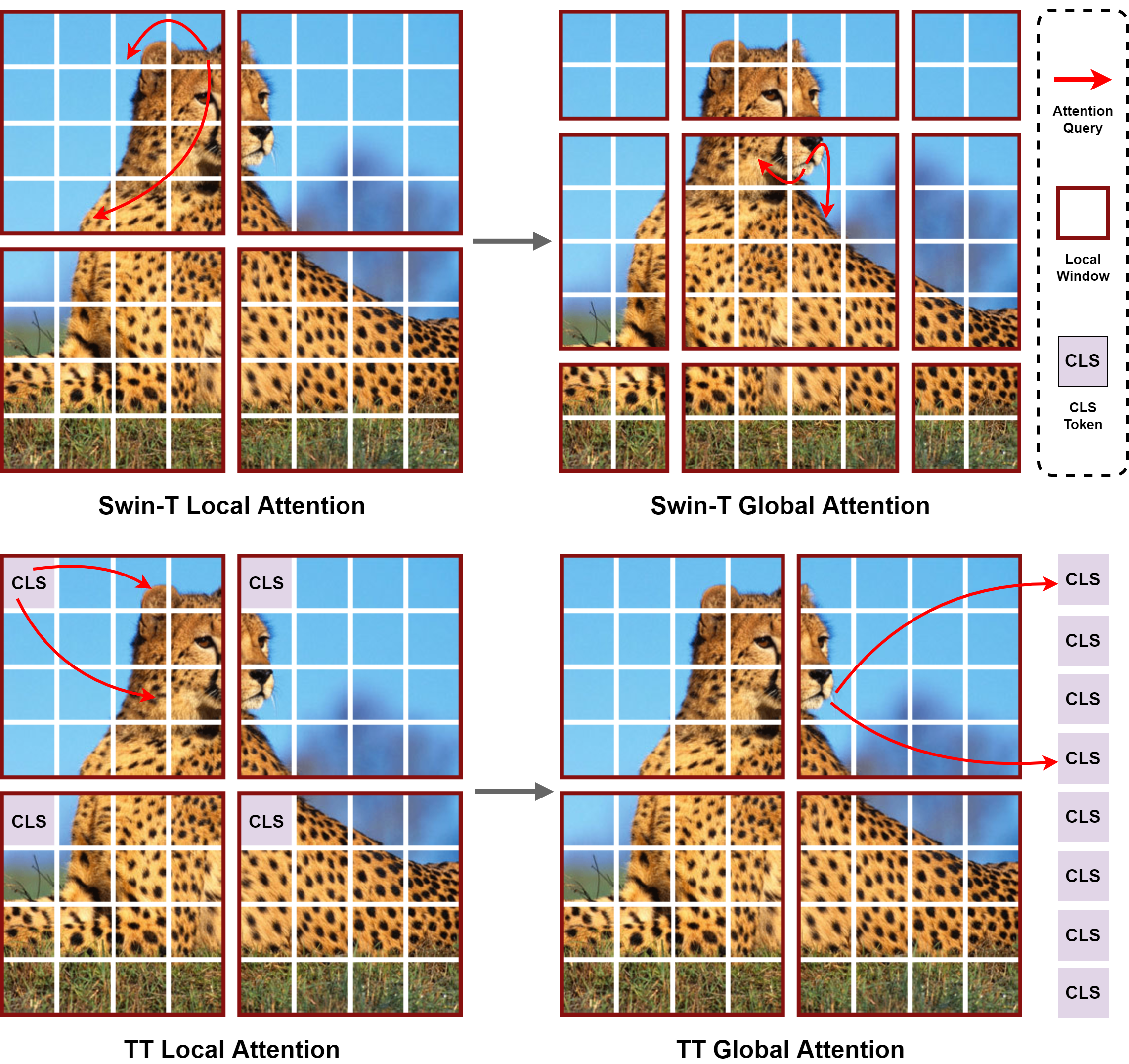}

   \caption{ Compare Swin-T's token interaction with TT. Swin-T enables long-range interaction between tokens through a window shift 	operation. 
   However, the size of the window and the shift scheme limit its ability to interact at a distance. 
   Our TT establishes efficient long-range dependency modeling while preserving the benefits of Swin-T by making CLS tokens capture window global information and making the tokens in each window interact with CLS tokens.}
   \label{fig1}
\end{figure}

Some recent work has begun to try to address this issue. Wang et al \cite{wang2021pyramid}. introduced the pyramidal architecture of CNN into ViT to alleviate the problem of huge computation of ViT by reducing the resolution in a hierarchical way. 
Yin et al \cite{yin2022vit}. studied unimportant tokens in ViT and greatly reduced the throughput of the model by suspending unimportant tokens. 
Although these works have made good progress, the Swin Transformer (Swin-T) \cite{liu2021swin} proposed by Liu et al. achieves a better tradeoff between accuracy and efficiency through window attention as well as window shift operations. 
However, the window size and window shift operations limit the long-range dependency modeling of the transformer. This in turn impairs the model performance.

In order to address the above issue, a QnA \cite{arar2022learned} layer is designed by introducing a trainable cross-window shared Query to further improve the expression of window-based transformer attention mechanisms by means of displacement invariance. 
Xia et al. proposed Deformable Attention Transformer (DAT)\cite{xia2022vision}, which indirectly improves the model modeling capability by selecting more meaningful key-value pairs based on the data. 
Yang et al.\cite{yang2021focal} effectively captured short- and long-range visual dependencies by investigating new attention mechanisms that combine fine-grained local and coarse-grained global interactions.

While these efforts have slightly improved window-based transformer modeling capabilities, they require the addition of additional layers or large overhead operations. 
So in this paper, we propose a novel Token Transformer (TT) to solve the above problem. The key innovation of TT is CLS Attention, which introduces a CLS token for each window. 
The success of ViT and MoCo v3\cite{chen2021empirical} has demonstrated that CLS tokens already have the effect of summarizing global information by interacting spatially with embedded tokens. 
Inspired by these works, we introduce a CLS token in each window to represent the information of each window by way of window attention. 
The CLS token containing the window information will be used as the query to interact spatially with the token between each window by cross-attention, thus enabling efficient long-range dependency modeling. 
Hence, we do not need to introduce additional layers or complex computational operations. \cref{fig1} shows the token interaction of TT compared with Swin-T. 
However, window-based transformers often use a layered design, which inevitably leads to a change in the number of windows and thus we need to create new CLS tokens when layering. 
In order to enable new CLS tokens to access the information contained in the previous CLS token, we design a Feature Inheritance Module (FIM) at each stage. 
It aggregates the previous stage CLS tokens by convolution and fuses the two stages of CLS tokens by spatial mixing. Besides, we design a simple and effective feedforward network in TT. 
We refer to the feedforward network in TT as Spatial-Channel Feedforward Network (SCFFN). 
In SCFFN, we perform token mixing not only on the spatial domain using Multi-Layer Perception (MLP), but also on the channel domain by convolution. 
This approach also introduces the inductive bias property of convolution for TT. And we do not increase the number of parameters of the feedforward network because we replaced the second layer of the conventional FFN network MLP with a 1x1 convolution. 
Such a feedforward network design is intuitive and simple, and helps us to enhance the ability to model the network over the channel domain.

We have conducted experiments to verify the effectiveness of TT. 
TT achieves competitive results on image classification task on ImageNet-1k \cite{deng2009imagenet}, semantic segmentation task on ADE20K \cite{zhou2019semantic}, and object detection task on CoCo\cite{lin2014microsoft} with fewer parameters.

In general, our main contributions are shown below:
\begin{itemize}
\item We designed a new long-distance dependency interaction called CLS Attention by introducing CLS tokens for each window. This interaction does not require the addition of additional layers or expensive computational operations, thus making the model lightweight and efficient. Based on this interaction, we have designed a novel transformer called TT, which has achieved competitive results in multiple visual tasks. 
\item To preserve the hierarchical design of window-based transformer, we designed the FIM for efficient message delivery of CLS tokens at different stages.
\item We designed a novel feedforward network SCFFN in TT. SCFFN enhances the modeling capability of TT simply and intuitively without introducing additional parameters by using spatial channel token mixture implemented by MLP and convolution, respectively.
\end{itemize}

\begin{figure*}[t]
  \centering
  \includegraphics[width=1\linewidth]{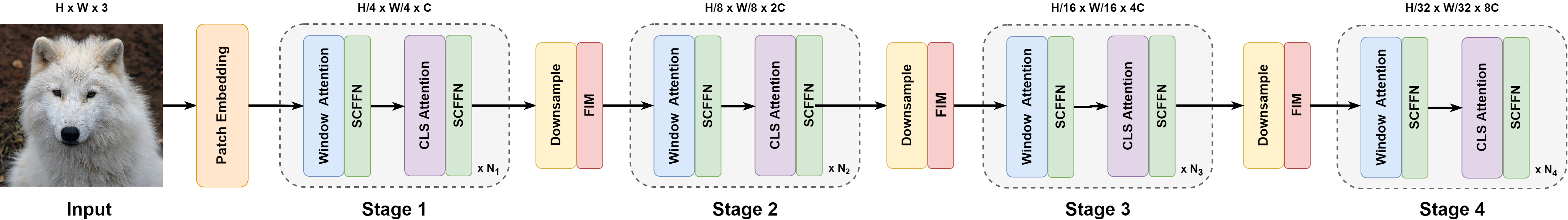}

   \caption{TT's overall pipeline. $N_1$ to $N_4$ represent the number of token transformer modules in each stage, respectively.}
   \label{fig2}
\end{figure*}
\section{Related Work}

\subsection{ViT and its variants}
The emergence of ViT\cite{dosovitskiy2020image} has successfully migrated the mainstream architecture transformer in Natural Language Processing (NLP) to the field of computer vision. 
ViT and its improvements has successfully outperformed or rivaled CNNs in several vision tasks with its global modeling capabilities. 
Zhou et al.\cite{zhou2021deepvit} found that the transformer quickly saturates at deeper levels. 
Therefore, they proposed the Re-attention model in DeepViT to regenerate the attention map to enhance the diversity among layers with little computational cost. 
CrossViT\cite{chen2021crossvit} proposes a dual-branch transformer module to combine image patches of different sizes to produce stronger image features to improve the image classification performance of ViT. 
However, the secondary computational complexity associated with global modeling has affected the development of ViT. 
To solve this problem, Yuan et al.\cite{yuan2021tokens} proposed T2T by incorporating the experience of CNN architecture design. 
With its narrow and deep network design, T2T further improves the performance of ViT while reducing the number of ViT parameters. CCT\cite{hassani2021escaping} proposes seif-attention processing using convolution rather than direct image chunking. 
This allows CCT to have higher accuracy and fewer parameters. PiT\cite{heo2021rethinking} combines a pooling layer with a transformer architecture to reduce the model computation in a way that reduces the space size. 
CvT\cite{wu2021cvt} improves the performance and efficiency of ViT by introducing convolution into ViT to produce the best results of both designs. 
Wang et al.\cite{wang2021pyramid} further improved the efficiency of ViT by transforming it into a PVT with a pyramidal architecture. 
RegionViT \cite{chen2021regionvit} uses a novel Regional-to-Local Attention instead of Global Self-Attention in ViT to reduce the computational effort. 
AViT\cite{yin2022vit} introduces a pause probability for each layer of embedded tokens, improving ViT efficiency by reducing unimportant token interactions.

\subsection{Window-based transformer}
Swin transformer\cite{liu2021swin} succeeds in reducing the secondary computational complexity of ViT to linear with its window-based attention design and window shift operation. 
With these careful designs, Swin-T perfectly achieves the trade-off between accuracy and computational volume. 
Despite these excellent results, its long-range modeling capability is somewhat impaired by the limitations of the window size and shift operations. 
To address this issue, Dong et al.\cite{dong2022cswin} developed the Cross-Shaped Window self-attention mechanism for computing self-attention in parallel in the horizontal and vertical stripes that form the cross-shaped window. 
This design achieves a powerful modeling capability while limiting the computational cost. Xie et al.\cite{xia2022vision} proposed DAT to solve this problem by finding the most relevant key-value pairs based on the intrinsic relationship of the data. 
Focal Transformer\cite{yang2021focal} enhances network modeling capabilities by designing complex coarse-grained local and fine-grained global interaction mechanisms. 
Li et al.\cite{li2021local} proposed a multi-branch, multi-scale Self-Attention-aware approach to LG-Transformer indirectly enhancing the modeling capability of Swin-T. 
MOA-Transformer \cite{patel2022aggregating} designs multi-resolution overlapping attention modules to generate global features. 
Arar et al.\cite{arar2022learned} designed window-shared learnable query for window-based transformer to introduce convolutional locality and variability to improve the expressiveness of the attention mechanism.
Ding et al.\cite{ding2022davit} designed Channel Group Attention, which allows their DaViT to model the global space at the channel level with linear computational complexity. 
In addition Wang et al.\cite{wang2022convolutional}  introduced convolutional embeddings for each transformer module to improve the modeling ability of hierarchical vision transformers such as swin-T, Cswin-T. 
Inspired by the dilated convolution \cite{yu2017dilated}, Hassani et al.\cite{hassani2022dilated} proposed Dilated Neighborhood Attention, which addresses the lack of Swin's ability to model long-range dependencies by expanding the receptive field.

\section{Method}
\subsection{Preliminaries}
We first review the attention mechanism in swin transformer \cite{liu2021swin}. 
Swin-T mainly adopts window based multi-head self-attention (W-MSA) mechanism and shifted window based multi-head self-attention (SW-MSA) mechanism. 
Given an input image ${\rm{x}} \in {R^{H \times W \times 3}}$, the patch splitting module splits it into non-overlapping patches. 
The linear embedding layer will project them as sequential patches ${{\rm{x}}_p} \in {R^{N \times C}}$, 
where $H$ and $W$ are the height and width of the input image, $C = {{\rm{p}}^2}e$ , $p$ denotes the patch size, $e$ denotes the embedding dimension, and $N = HW/{{\rm{p}}^2}$ denotes the number of patches. 
Assume that each local window contains $M$ patches ${\rm{z}} \in {R^{M \times C}}$. W-MSA performs multi-headd attention on the patches in each window separately. The W-MSA with $N$ heads is specified as

\begin{equation}
  \begin{split}
    &q = {z}{W_q},k = {z}{W_k},v = {z}{W_v},\\
    &{o^{(n)}} = \theta ({q^{(n)}}{k^{(n)T}}/\sqrt d  + B){v^{(n)}},n = 1,...,N,\\
    &o = [{o^{(1)}},...,{o^{(N)}}]{W_o},
\end{split}
  \label{eq1}
\end{equation}

In \cref{eq1}, ${W_q},{W_k},{W_v},{W_o}$ represents the mapping matrix. $\theta ( \cdot )$ is the softmax function, 
$d = C/N$ is the embedding dimension of each header and $B \in {R^{N \times N}}$ is the relative position bias.
$\left[  \cdot  \right]$means Concat operation. 
Next, each window is shifted and then the W-MSA calculation is performed once again, resulting in a long-range dependency modeling, also known as SW-MSA. 
Overall, the attention mechanism of Swin-T is calculated as
\begin{equation}
    \begin{aligned}
    &{{\hat{o}}^{l}} = \text{W-MSA}\left( {\text{LN}\left( {{{o}^{l - 1}}} \right)} \right) + {o}^{l - 1},\\
    &{{o}^l} = \text{MLP}\left( {\text{LN}\left( {{{\hat{o}}^{l}}} \right)} \right) + {{\hat{o}}^{l}},\\
    &{{\hat{o}}^{l+1}} = \text{SW-MSA}\left( {\text{LN}\left( {{{o}^{l}}} \right)} \right) + {o}^{l},\\
    &{{o}^{l+1}} = \text{MLP}\left( {\text{LN}\left( {{{\hat{o}}^{l+1}}} \right)} \right) + {{\hat{o}}^{l+1}},
  \end{aligned}
  \label{eq2}
\end{equation}
where ${\rm{LN}}\left(\cdot\right)$ denotes Layer Normalization.
\begin{figure*}[t]
  \centering
  \includegraphics[width=0.9\linewidth]{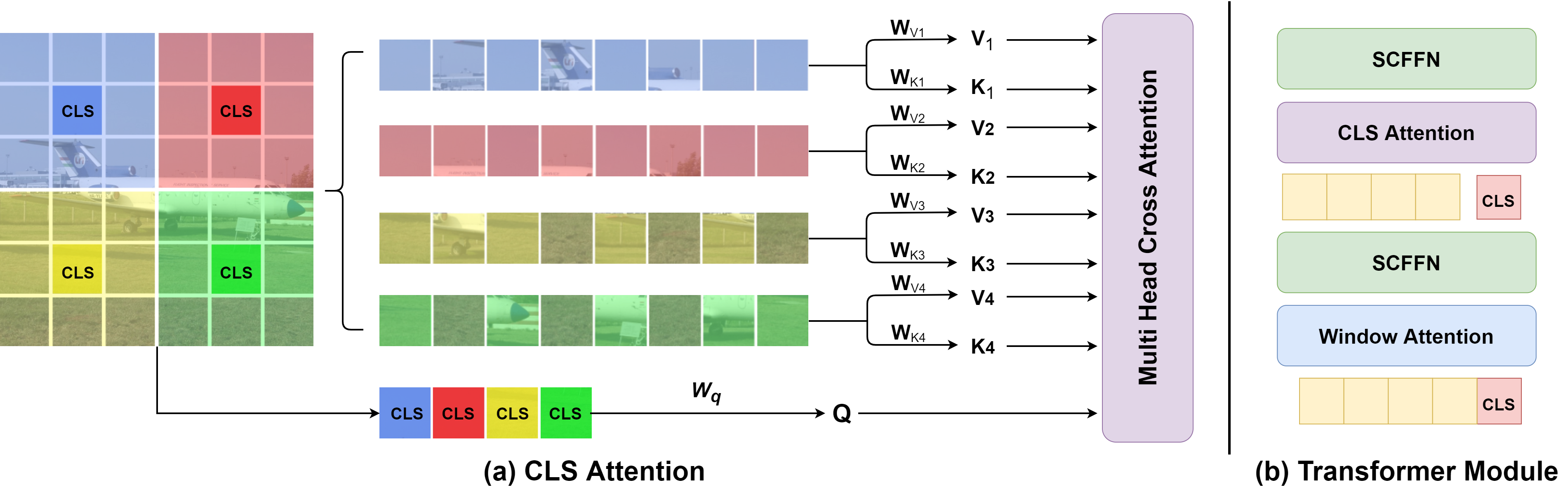}
   \caption{ (a) illustrates the CLS Attention mechanism. The CLS token first captures local window information through window attention. 
   The area with the same color as the CLS token indicates the receptive field of the CLS token. 
   Then, all CLS tokens are projected as queries, and the tokens in each window are projected as keys and values respectively. All keys and values share the same query. 
   Eventually, long-range dependency modeling is achieved by multi head cross attention. (b) describes the internal construction of the token transformer module.}
   \label{fig3}
\end{figure*}

\subsection{Overall Architecture}

\cref{fig2} illustrates the overall architecture of TT. TT mainly consists of patch embed layer, token transformer module and feature inheritance module. 
The model architecture remains a hierarchical design. Each stage reduces the feature map resolution by a factor of two. 
For the input image x, we divide it into non-overlapping patches and project these patches by a patch embedding layer composed of convolution. 
We then assign the patch to multiple non-overlapping local windows and add a CLS token to each window. 
Finally, we input the tokens from each window into the token transformer module for local and global modeling. 
In addition to downsampling during the layering phase, we also design the feature inheritance module for passing information from the previous phase window to the new CLS token.

\subsection{Token Transformer Module}

The token transformer module mainly consists of local window attention and global attention based on CLS tokens (CLS Attention). 
Besides, we design a Spatial-Channel Feedforward Network in the token transformer module which enhances network modeling capabilities in a simple and intuitive way. 
We show the CLS Attention of the token transformer module in \cref{fig3}.

\paragraph{CLS Attention} For token ${z} \in {R^{M \times C}}$ in each window, we add a CLS token $ \in {R^{1 \times C}}$ to it to get ${\hat{z}} \in {R^{(M + 1) \times C}}$.
For ${\hat{z}}$,we perform local modeling of window attention and make the CLS token aware of the local information of the window. The process is computing according to:


\begin{equation}
\begin{aligned}
      &{\hat{z}} = [\text{cls,z}],\\
      &{\hat{z}} = \text{W-MSA}\left( {\text{LN}\left( {{{\hat{z}}}} \right)} \right) + {\hat{{z}}},
\end{aligned}
\label{eq3}
\end{equation}
In the next stage, we integrate CLS tokens from all windows and use them as queries for long-range dependency modeling. 
The CLS token will interact with the embedded tokens $z$ in each local window, which will be used as key and value. 
The whole interaction process is realized through the cross-attention mechanism. \cref{eq4} illustrates this.

\begin{equation}
  \begin{aligned}
      &{{z}_q} = [cl{s^{(1)}},...,cl{s^{(T)}}],\\
      &q = {z_q}{W_q},k = z{W_k},v = z{W_v},\\
      &{o^{(n)}} = \theta ({q^{(n)}}{k^{(n)T}}/\sqrt d  + B){v^{(n)}},n = 1,...,N,\\
      &o = [{o^{(1)}},...,{o^{(N)}}]{W_o},
  \end{aligned}
  \label{eq4}
  \end{equation}
  where $T$ is the number of windows, ${W_{\rm{q}}} \in {R^{C \times C}}$ is the mapping matrix of CLS tokens, and ${W_{\rm{k}}},{W_v} \in {R^{C \times C}}$ are the projection matrices of the embedded tokens in the windows. 
  By cross-attention based on CLS tokens and embedded tokens in local windows, we have successfully implemented the modeling of long-range dependencies. 
  We provide the pseudo-code for the pytorch version of TT implementing long-range interaction in \cref{alg1}.

  \begin{algorithm}[h]
    \caption{CLS Attention: PyTorch-like Pseudocode}
    \algcomment{
    \textbf{Notes}:\texttt{num\_windows} indicates the number of windows.\texttt{rearrange()} is a shape change operation.
    }
    \definecolor{codeblue}{rgb}{0.25,0.5,0.5}
    \definecolor{codekw}{rgb}{0.85, 0.18, 0.50}
    \label{alg1}
    \begin{lstlisting}[language=python]
  class WindowAttentionGlobal(nn.Module):
      def __init__(
              self, dim, window_size, num_windows, attn_drop=0., proj_drop=0.):
          super().__init__()
          self.q = nn.Linear(dim, all_head_dim, bias=False)
          self.k = nn.Linear(dim, all_head_dim, bias=False)
          self.v = nn.Linear(dim, all_head_dim, bias=False)
          self.attn_drop = nn.Dropout(attn_drop)
          self.proj = nn.Linear(all_head_dim, dim)
          self.proj_drop = nn.Dropout(proj_drop)

      def forward(self, x):
          B_, N, C = x.shape
          B = int(B_ // self.num_windows)
          cls_token, token = x[:, :1], x[:, 1:]
          cls_token = rearrange(cls_token, (b n) d c -> b n d c, b=B).repeat(1, 1, self.num_windows, 1)
          cls_token = rearrange(cls_token, b n d c -> (b d) n c)
          k = v = token
          q = cls_token
          N_k, N_v = k.shape[1], v.shape[1]
          q = F.linear(input=q, weight=self.q.weight)
          q = q.reshape(B, N, 1, self.num_heads, -1).permute(2, 0, 3, 1, 4).squeeze(0)
          k = F.linear(input=k, weight=self.k.weight)
          k = k.reshape(B, N_k, 1, self.num_heads, -1).permute(2, 0, 3, 1, 4).squeeze(0)
          v = F.linear(input=v, weight=self.v.weight)
          v = v.reshape(B, N_v, 1, self.num_heads, -1).permute(2, 0, 3, 1, 4).squeeze(0)
          attn = self.attn_drop((q @ k.transpose(-2, -1)).softmax(dim=-1))
          x = (attn @ v).transpose(1, 2).reshape(B, N, -1)
          x = self.proj_drop(self.proj(x))
          return x
    \end{lstlisting}
    \end{algorithm}

  \paragraph{SCFFN.} We propose a novel feedforward network without introducing additional parameters. 
  In addition to retaining the spatial mixing function of the feedforward network, we have added a channel mixing mechanism to enhance the network modeling capability. 

  Given the embedding token $z$ and the CLS tokens in all windows, SCFFN first performs spatial mixing on them by MLP and then transposes the dimensions to perform channel mixing. 
  Generally, SCFFN are computed as

  \begin{equation}
    \begin{aligned}
        &{z} = \text{MLP}\left( \text{LN}\left(z \right) \right) +  z,\\
        &z = \text{Conv}\left(\text{LN}\left(\text{trans}\left(z\right)\right)\right) + \text{trans}\left(z\right), \\
        &z = \text{trans}\left(z\right), \\
        &\text{cls} = \text{SCFFN}\left(\text{cls}\right),\\
        &o = \left[\text{cls,z }\right],
    \end{aligned}
    \label{eq5}
    \end{equation}

    where $trans( \cdot )$ represents the dimension transpose operation, $Conv( \cdot )$ is the convolution operation with a convolution kernel of 1. 
    Compared with the conventional FFN, we did not increase the number of parameters since we only replaced the MLP in its second layer using 1x1 convolution. 
    However, this intuitive and simple feedforward design adds the benefits of convolutional inductive bias and channel modeling to TT, allowing for more sophisticated modeling.

\subsection{Feature Inheritance Module}

\begin{figure}[h]
  \centering
  \includegraphics[width=1\linewidth]{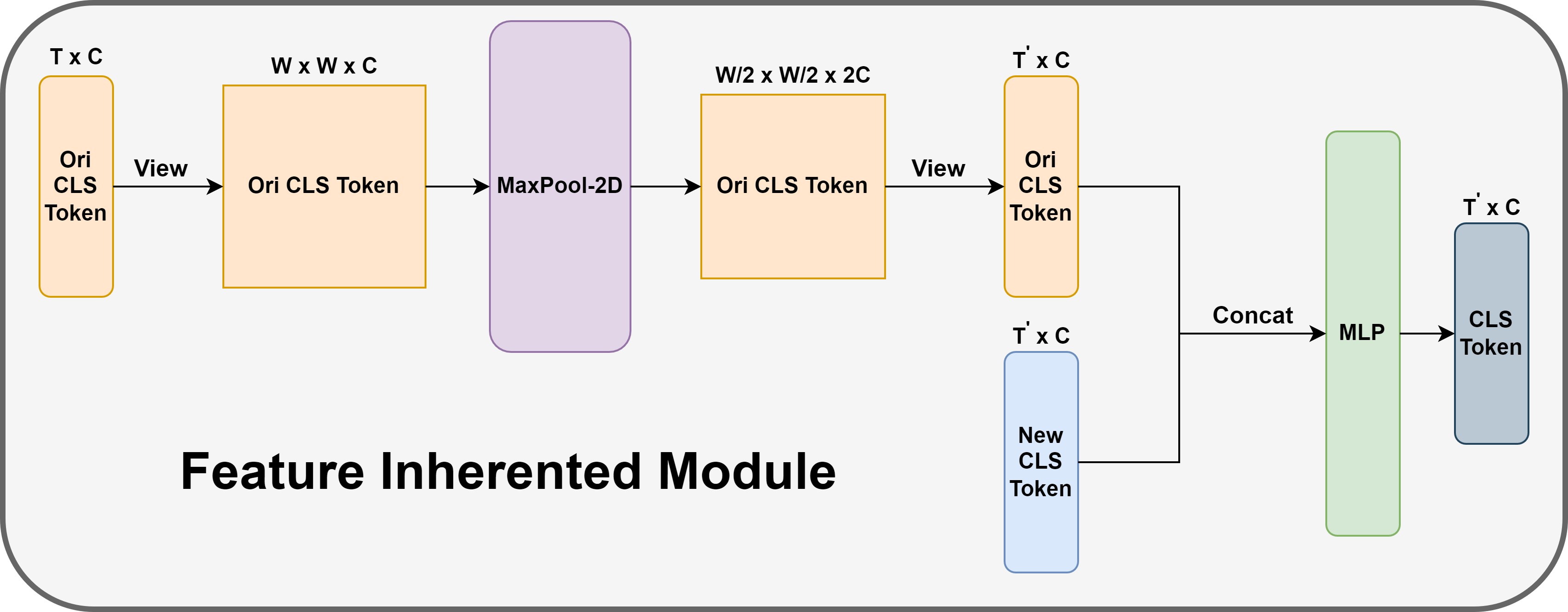}
   \caption{FIM architecture. We first divide the tokens into horizontal and vertical axis CLS tokens. 
   Next, we downsample and integrate the horizontal and vertical axis CLS tokens through the max-pooling layer. 
   We then merge the horizontal and vertical axis 	CLS tokens and combine them with the new CLS tokens by channel dimension. 
   Ultimately we transfer information between the 	origin CLS tokens and new CLS tokens through the MLP layer.}
   \label{fig4}
\end{figure}

Since window-based transformers usually use a hierarchical design, this inevitably leads to a reduction in the number of windows and thus we need to design new CLS tokens at each stage. 
However, the new CLS token is missing the local window information of the previous resolution feature map and thus does not facilitate the use of multi-scale features. 
To solve this issue, we designed feature inheritance modules for passing CLS token information in each phase of TT. \cref{fig4} shows the detailed flow of the FIM.

Specifically, supposing that all CLS tokens are a $T \times C$ matrix at a certain stage, FIM will first extend $T$ to $H,W = \sqrt T $ . $H$ and $W$ denote the number of horizontal and vertical CLS tokens in the feature map, respectively. 
Next, we perform a down-sampling operation on the CLS tokens $ \in {R^{H \times W \times C}}$ by a $3 \times 3$ convolutional layer with stride of $2$ and padding of $1$. 
This step successfully integrates CLS tokens of the feature map in both horizontal and vertical directions and expands the embedding dimension of CLS tokens. 
Then FIM will transform the previous CLS tokens $ \in {R^{(H/2) \times (W/2) \times 2C}}$ into the shape of the new CLS tokens and merge them by channel dimension. 
Finally, we use the projection matrix ${W_{\rm{c}}} \in {R^{4C \times 2C}}$ to integrate the CLS tokens for the two stages. \cref{alg2} illustrates the pseudo-code of FIM based on Pytorch.
\begin{algorithm}[h]
  \caption{FIM: PyTorch-like Pseudocode}
  \algcomment{
  }
  \definecolor{codeblue}{rgb}{0.25,0.5,0.5}
  \definecolor{codekw}{rgb}{0.85, 0.18, 0.50}
  \label{alg2}
  \begin{lstlisting}[language=python]
class FeatureInheritanceModule(nn.Module):
    def __init__(self, dim, kernel_size=3, stride=2, pad=1):
        super(FeatureInheritanceModule, self).__init__()
        self.maxpool = nn.MaxPool2d(kernel_size=kernel_size, stride=stride, padding=pad)
        self.cls_mlp = nn.Linear(in_features=int((dim // 2) * 3), out_features=dim, bias=False)

    def forward(self, ori_cls, new_cls):
        CB, CN, CC = new_cls.shape
        B, N, C = ori_cls.shape
        batch = CB
        Window = int(B // batch)
        W = int(Window ** 0.5)
        ori_cls = ori_cls.view(batch, C, W, W)
        ori_cls = self.maxpool(ori_cls)
        ori_cls = ori_cls.view(CB, CN, CC // 2)
        cls_token = torch.cat([ori_cls, new_cls], dim=-1)
        cls_token = self.cls_mlp(cls_token)
        return cls_token
  \end{lstlisting}
  \end{algorithm}

\subsection{Architecture Variants}

We designed three different versions of Token Transformer according to the parameter size: TT-T, TT-S and TT-B. 
The hyperparameters and network configurations for each version of Token Transformer are listed in \cref{tab1}.

\begin{table}[h]
  \begin{center}
    \footnotesize
  \setlength{\tabcolsep}{0.01mm}{
  \renewcommand\arraystretch{1.0}
  \begin{tabular}{c|c|c|c}
  \thickhline
  \multicolumn{4}{c}{\textbf{\ourmethod{} Architectures}} \\
   & \ourmethod-T & \ourmethod-S & \ourmethod-B \\
  \hline
  \multirow{2}{*}{
  $\begin{matrix}\text{Stage 1}\\(\text{49}\!\times{}\!\text{49})\end{matrix}$
  } 
  & $N_1\!=\!3,C\!=\!64$ & $N_1\!=\!4,C\!=\!96$ & $N_1\!=\!5,C\!=\!128$ \\
  & $W\!=\!7,H\!=\!2$ & $W\!=\!7,H\!=\!3$ & $W\!=\!7,H\!=\!4$ \\
  \hline
  \multirow{2}{*}{
  $\begin{matrix}\text{Stage 2}\\(\text{25}\!\times{}\!\text{25})\end{matrix}$
  } 
  & $N_2\!=\!4,C\!=\!128$ & $N_2\!=\!5,C\!=\!192$ & $N_2\!=\!6,C\!=\!256$ \\
  & $W\!=\!5,H\!=\!4$ & $W\!=\!5,H\!=\!6$ & $W\!=\!5,H\!=\!8$ \\
  \hline
  \multirow{2}{*}{
  $\begin{matrix}\text{Stage 3}\\(\text{16}\!\times{}\!\text{16})\end{matrix}$
  } 
  & $N_3\!=\!19,C\!=\!256$ & $N_3\!=\!21,C\!=\!384$ & $N_3\!=\!22,C\!=\!512$ \\
  & $W\!=\!4,H\!=\!8$ & $W\!=\!4,H\!=\!12$ & $W\!=\!4,H\!=\!16$ \\
  \hline
  \multirow{2}{*}{
  $\begin{matrix}\text{Stage 4}\\(\text{9}\!\times{}\!\text{9})\end{matrix}$
  } 
  & $N_4\!=\!5,C\!=\!512$ & $N_4\!=\!5,C\!=\!768$ & $N_4\!=\!5,C\!=\!1024$ \\
  & $W\!=\!3,H\!=\!16$ & $W\!=\!3,H\!=\!24$ & $W\!=\!3,H\!=\!32$ \\
  \thickhline
  \end{tabular}}
  \end{center}
  \vskip -0.1in
  \caption{Details of the configuration of TT and its variants. N represents the number of modules, C is the embedding dimension, W denotes the window size, and H refers to the number of attention heads.}
  \label{tab1}
  \vskip -0.2in
  \end{table}

\section{Experiments}

We perform image classification experiments with Token Transformer on ImageNet-1k \cite{deng2009imagenet}, a object detection task on CoCo \cite{lin2014microsoft}, and a semantic segmentation performance test on ADE20K \cite{zhou2019semantic}. 
Next, we verify the effectiveness of each important component of TT by ablation experiments on ImageNet-1k.

\subsection{Image Classification on ImageNet-1K}

\begin{table}[h]
    \centering
    \footnotesize
    \setlength{\tabcolsep}{1.0mm}
    \begin{tabular}{l|c|c|c|c}
      
          Method &  Resolution & \#Params & \#Flops &Top-1 \\ 
          \whline
          Swin-T\cite{liu2021swin} & $224^2$ & 29M &4.5G& 81.3\\
          ConvNeXt-T\cite{liu2022convnet} & $224^2$ & 29M &4.5G& 82.1\\
          Focal-T\cite{yang2021focal} &$224^2$ & 29M &4.9G& 82.2\\
          PVT-S\cite{wang2021pyramid} & $224^2$ & 24M &3.8G& 79.8\\
          ViT-Small\cite{dosovitskiy2020image}& $224^2$ & 24M &5.1G& 82.0\\
          DAT-T\cite{xia2022vision} & $224^2$ & 29M &4.6G& 82.0\\
          LG-T\cite{li2021local}& $224^2$ & 32M &4.8G& 82.1\\
          QnA-Tiny\cite{arar2022learned}& $224^2$  & 16M &2.5G& 81.7\\
          \rowcolor{mygray}
          TT-T & $224^2$ & 25M& 3.9G & 82.5\\
          \hline
          Swin-S\cite{liu2021swin} & $224^2$ & 50M &8.8G& 83.0\\
          ConvNeXt-S\cite{liu2022convnet} & $224^2$ & 50M &8.7G& 83.1\\
          Focal-S\cite{yang2021focal} & $224^2$ & 51M &9.1G& 83.5\\
          PVT-M\cite{wang2021pyramid}& $224^2$ & 44M &6.9G& 81.2\\
          ViT-Medium\cite{dosovitskiy2020image} & $224^2$ & 39M &9.1G& 83.3\\
          DAT-S\cite{xia2022vision}& $224^2$ & 50M &9.0G& 83.7\\
          LG-S\cite{li2021local}& $224^2$ & 61M &9.4G& 83.3\\
          QnA-Small\cite{arar2022learned}& $224^2$  & 25M &4.4G& 83.2\\	
          \rowcolor{mygray}
          TT-S & $224^2$ & 47M & 7.7G & 83.7\\
          \hline
          Swin-B\cite{liu2021swin} & $224^2$ & 88M &15.5G& 83.5\\
          ConvNeXt-B\cite{liu2022convnet}& $224^2$ & 89M &15.4G& 83.8\\
          Focal-B\cite{yang2021focal} & $224^2$ & 89M &16.0G&83.8 \\
          PVT-L\cite{wang2021pyramid}& $224^2$ & 61M &9.8G& 81.7\\
          ViT-Base/16 \cite{dosovitskiy2020image} & $224^2$ & 86M &17.6G& 77.9\\
          DAT-B\cite{xia2022vision}/16& $224^2$ & 88M &15.8G& 84.0\\
          QnA-Base\cite{arar2022learned} & $224^2$ & 56M &9.7G &83.7\\
          \rowcolor{mygray}
          TT-B& $224^2$ & 86M &14.6G& 84.2\\ 

      \end{tabular}
    \caption{TT is compared with other state-of-the-art algorithms on ImageNet-1k for top-1 accuracy, parametric number size.}
\label{tab2}
\end{table}

\paragraph{Settings.}

We pre-train and fine-tune TT on the ImageNet-1k dataset. The ImageNet-1k dataset consists of a training set of 1.28 million images and a validation set of 500,000 images. 
The dataset is divided into a total of 1000 categories. 
We test the performance of our three variants on ImageNet-1k and compared it with other transformer models.

All models are performed on $224 \times 224$ resolution images. We use the AdamW optimizer with an initial learning rate of 1e-4 for gradient updating. 
In addition, the model is trained with a total of 350 epochs and a weight decay of 0.05. 
The learning rate strategy employs a cosine decay scheduler with 20 warm-up and cool-down cycles. Moreover, we use most of the data enhancement strategies. 
All models are implemented in Nvidia RTX 3090.

\paragraph{Results.} We provide the experimental results of image classification on ImageNet-1k in \cref{tab2}. We compare with current ViT, CNN, and window-based models. 
It is worth mentioning that our TT and its various variants achieve state-of-the-art classification performance with a small number of parameters and less computational complexity.

\subsection{Object Detection on COCO}

\begin{table}[h]
  \begin{center}
  \footnotesize
  \setlength{\tabcolsep}{1.0mm}{
  \addtolength{\tabcolsep}{-2.5pt}
  \begin{tabular}{c|c|c|c|c|c|c|c|c}
  Method & \#Params & \#Flops& $AP^b$ & $AP^b_{50}$ & $AP^b_{75}$ & $AP^m$ & $AP^m_{50}$ & $AP^m_{75}$ \\
  \whline
  Swin-T\cite{liu2021swin} & 48M&267G & 46.0	& 68.1 & 50.3	& 41.6 & 65.1	& 44.9\\
  ConvNeXt-T\cite{liu2022convnet} & 48M&262G & 46.2	& 67.9 & 50.8	& 41.7 & 65.0	& 44.9\\
  DAT-T\cite{xia2022vision} & 48M & 272G&47.1 & 69.2	& 51.6 & 42.4	& 66.1 & 45.5\\
  \rowcolor{mygray}
  TT-T & 44M&245G & 47.2	& 69.2 & 51.7 &	42.6 & 66.2	& 45.6\\
  \hline
  Swin-S\cite{liu2021swin} & 69M&359G & 48.5	& 70.2 & 53.5	& 43.3 & 67.3	& 46.6\\
  DAT-S\cite{xia2022vision}  & 69M & 378G&49.0	& 70.9 & 53.8	& 44.0 & 68.0	& 47.5 \\
  \rowcolor{mygray}
  TT-S & 66M &343G& 48.9 & 70.7	& 53.7 & 44.1	& 68.0 & 47.4\\
  \end{tabular}
  }
  \end{center}\vspace{-4mm}
  \caption{Comparison of TT with other models for object detection results on the CoCo dataset.}
  \label{tab3}\vspace{-2mm}
  \end{table}

\paragraph{Settings.} We use the CoCo dataset to evaluate the object detection performance of TT. 
The CoCo dataset consists of a training set of 118k images and a validation set of 5k images. 
We use TT and its variants trained on ImageNet-1k as Mask R-CNN \cite{he2017mask} 3 x schedule backbone. 
We use the training settings of Swin-T for object detection training on CoCo.

\paragraph{Results.}\cref{tab3} shows the target detection results of TT and its three variants as well as several comparison algorithms on the CoCo dataset. 
We use Swin-T as a baseline and compare TT with various SOTA object detection algorithms such as CNN-based ConvNext \cite{liu2022convnet} and window-based DAT \cite{xia2022vision}. 
The experiments show that TT and its various variants demonstrate competitive results for object detection on the CoCo dataset.

\subsection{Semantic Segmentation on ADE20K}
\begin{table}[h]
  \begin{center}
  \footnotesize
  \setlength{\tabcolsep}{1.5mm}{
  \addtolength{\tabcolsep}{-2.5pt}
  \begin{tabular}{c|c|c|c|c}
  Backbone & Method &  \#Params & \#Flops& mIoU\\
  \whline
  Swin-T\cite{liu2021swin}& UperNet & 60M	&945G& 44.5  \\
  DAT-T\cite{xia2022vision} & UperNet&  60M &957G&	45.54    \\
  DeiT-Small/16\cite{touvron2021training}  &UperNet  &  52M	&1099G& 44.0  \\
  Focal-T\cite{yang2021focal}      &UperNet    & 62M&998G	& 45.8 \\
  \rowcolor{mygray}
  TT-T & UperNet &  56M &883G&	46.3  \\
  \hline
  Swin-S\cite{liu2021swin} & UperNet &  81M	&1038G& 47.6  \\
  DAT-S\cite{xia2022vision} & UperNet &  81M&1079G	& 48.31 \\
  Focal-S\cite{yang2021focal} & UperNet &  85M &1130G&	48.0 \\
  \rowcolor{mygray}
  TT-S & UperNet &  78M&1006G & 48.0   \\
  \hline
  Swin-B\cite{liu2021swin} & UperNet & 121M&1188G	& 48.1  \\
  DAT-B\cite{xia2022vision}&UperNet       & 121M &1212G&	49.38  \\
  Focal-B\cite{yang2021focal} & UperNet &  126M &1354G&	49.0 \\
  \rowcolor{mygray}
  TT-B & UperNet&  117M &1151G&	48.8 \\
  \end{tabular}
  }
  \end{center}\vspace{-2mm}
  \caption{Comparison of TT with other models for semantic segmentation results on the ADE20K dataset.}
  \label{tab4}\vspace{-6mm}
  \end{table}

\paragraph{Settings.} ADE20K is an accepted standard dataset for semantic segmentation tasks. It contains 20K training images and 2K validation images. 
In the semantic segmentation task, we still use the pre-training weights of TT on ImageNet-1k, and use UperNet \cite{xiao2018unified} as the segmentation head. 
For a fair comparison, we still keep the same training settings as Swin-T in the semantic segmentation task.

\paragraph{Results.} We present the results of TT compared with various SOTA models in \cref{tab4}. It contains the number of model parameters and the mIoU metric for the semantic segmentation task. 
We still choose Swin-T as the baseline and select DeiT, Focal and DAT for comparison. 
Compared with other models, TT still achieves competitive results with fewer parameters, which demonstrates the effectiveness of TT in the semantic segmentation task.

\subsection{Ablation Study}
In this section, we verify their validity by ablating important components of TT through image classification experiments on ImageNet-1k. We next report their ablation results in detail.



\begin{figure}[h]
  \begin{minipage}[t]{0.5\linewidth}
    \centering
    \includegraphics[scale=0.5]{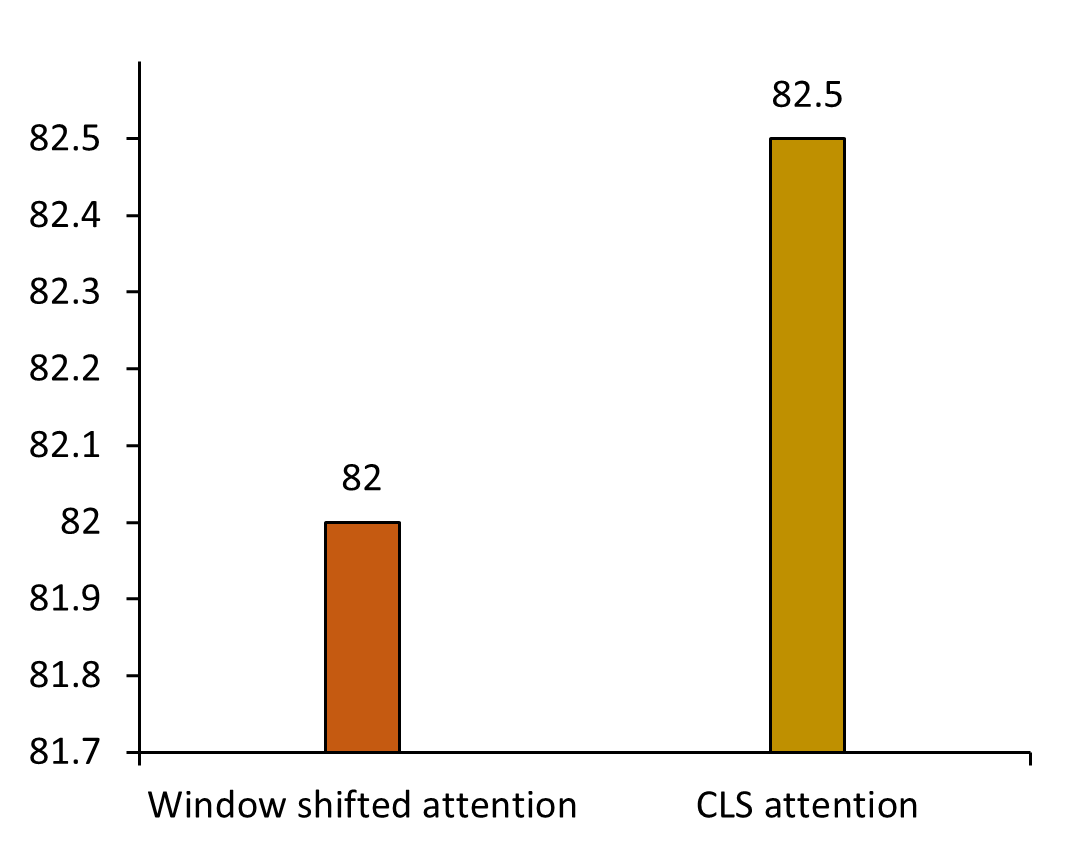}
    \caption{CLS Attention ablation results.}
    \label{fig5}
  \end{minipage}%
  \begin{minipage}[t]{0.5\linewidth}
    \centering
    \includegraphics[scale=0.5]{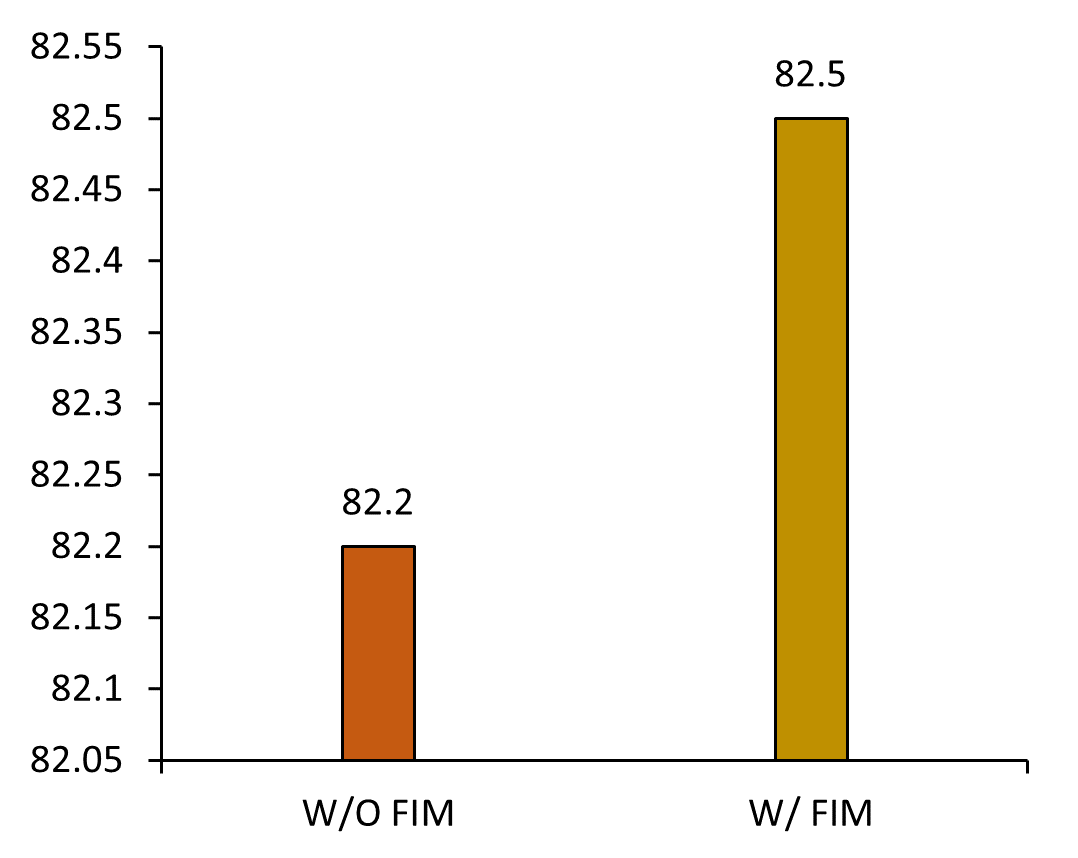}
    \caption{FIM ablation results.}
    \label{fig6}
  \end{minipage}
\end{figure}

\paragraph{CLS Attention.} We first verify the effectiveness of CLS Attention based on CLS tokens. 
Specifically, while we add CLS tokens to each local window, we eliminate the use of cross-attention for remote interaction between CLS tokens and embedded tokens in the local window. 
Instead, we use a shift window approach for long-range dependency modeling. 
As shown in \cref{tab5}, the Top-1 accuracy of image classification on ImageNet-1k with TT using shift operation decreased by 0.5 compared to the remote attention mechanism based on CLS token. 
This demonstrates the effectiveness of our proposed Long-range attention mechanism based on CLS token to capture global information.

\paragraph{Spatial-channel feedforward network.} For the ablation experiments of the spatial-channel feedforward network, we investigate the effectiveness of SCFFN acting on CLS tokens and embedded tokens. 
Specifically, we designed three classification experiments for testing using conventional feedforward networks instead of SCFFN without considering the number of parameters. 
If only SCFFN is used for CLS tokens and FFN is used for embedded tokens, the top-1 accuracy of TT-T decreases by 0.2. 
If only FFN is used for CLS tokens and SCFFN is used for embedded tokens, the classification experiment of TT-T crashes. 
If FFN is used for both CLS token makes and embedded tokens, the top-1 accuracy of TT-T decreases by 0.3. The specific results are detailed in \cref{fig5}. 
This indicates that SCFFN is more useful for CLS tokens and it enhances the modeling capability of TT more than the normal feedforward network.

 \begin{table}[h]
    \footnotesize
    \centering
    \begin{tabular}{c|cc|c}
      \toprule[1pt]
      Backbone & CLS token & Embed token & Top-1 acc \\
      \hline
      \multirow{4}[2]{*}{TT-T} & FFN   & FFN   & 82.2\gbf{-0.3} \\
            & SCFFN & FFN   & 82.3\gbf{-0.2} \\
            & FFN   & SCFFN & - \\
            \cmidrule{2-4}          
            & SCFFN & SCFFN & 82.5 \\
  \toprule[1pt]
    \end{tabular}
    \caption{Ablation studies of SCFFN on TT-T.}
    \label{tab5}
  \end{table}
  
\paragraph{Feature Inheritance Module.} Finally, we investigate whether the messaging of CLS tokens at different stages is effective. 
Specifically, we remove the feature inheritance modules in different phases and provide only new CLS tokens for each phase. 
\cref{fig6} shows the ablation results of FIM for image classification on ImageNet-1k. The results show that removing FIM decreases the Top-1 accuracy by 0.3. 
This result illustrates the importance of FIM for TT to model long-range dependencies using multi-scale features.

\subsection{Visualization}
  \begin{figure*}[h]
  \centering
  \includegraphics[width=1\linewidth]{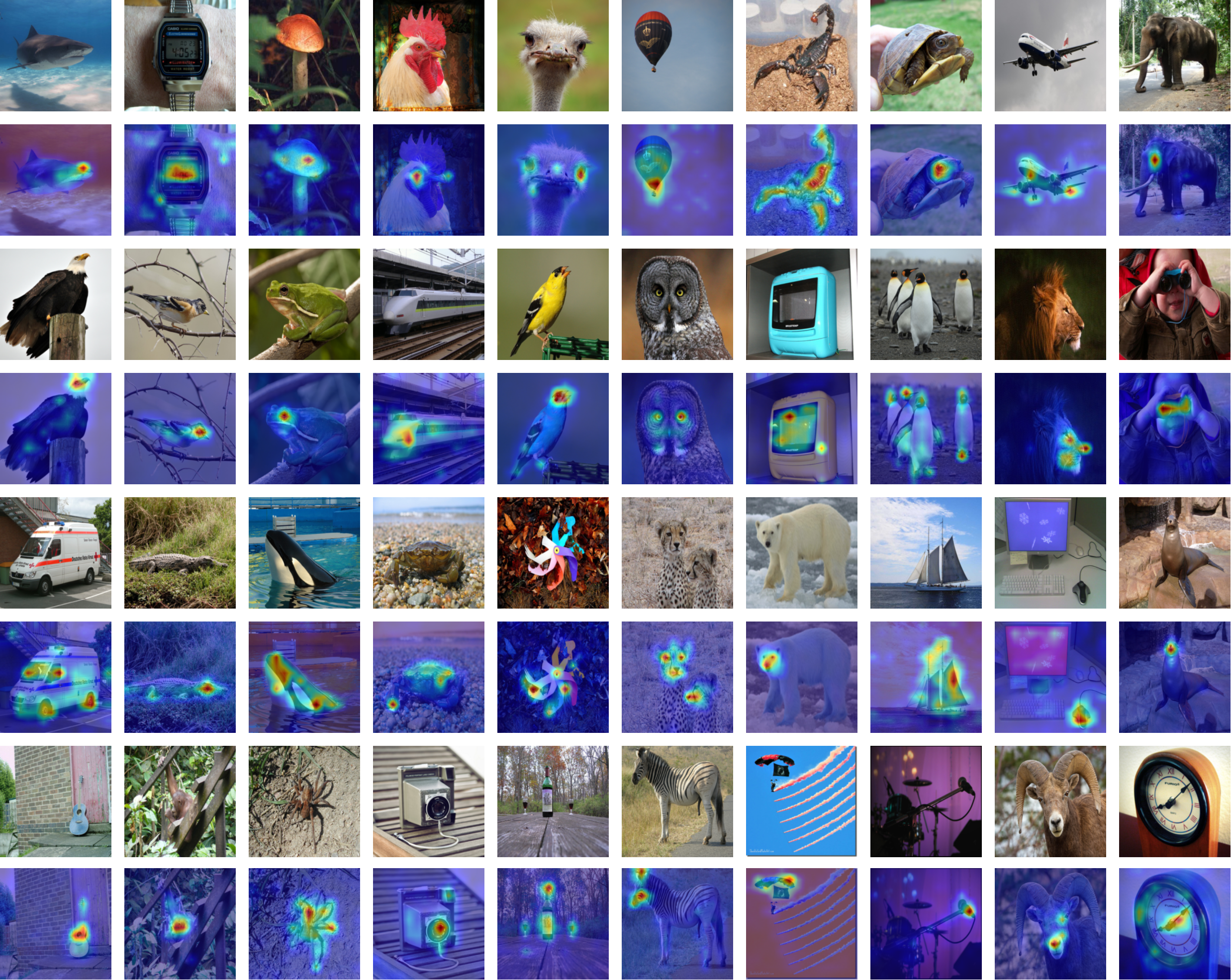}
   \caption{Attention visualization results of TT-T on ImageNet-1k dataset. We use the TT-T last layer attention map for visualization.}
   \label{fig7}
\end{figure*}
To demonstrate the performance of CLS Attention, we show the results of TT last layer attention visualization on the ImageNet-1k dataset. 
\cref{fig7} shows the TT focused areas of interest in various types of images.

\section{Conclusion}
In this paper we propose a novel token transformer (TT), which is a simpler and more efficient one and can be used for multiple vision tasks. 
The core mechanism of TT is a long-range attention mechanism based on CLS tokens. 
It implements more efficient modeling of long-range dependencies without introducing additional layers or complex computational operations. 
In addition, the simple and intuitive design of SCFFN helps TT to enhance its modeling capabilities while handling two different nature tokens at the same time. 
In order to solve the problem that TT hierarchical design leads to the next stage CLS token cannot access the window information of the previous stage, we design FIM in each stage of TT. 
Ablation experiments have verified its effectiveness. The proposed TT achieves competitive results in several vision tasks. 
We hope that our work will inspire the design of more efficient and simpler ways to model long range dependencies.

\section*{Acknowledge} This work was supported by Public-welfare Technology Application Research of Zhejiang Province in China under Grant LGG22F020032, and Key Research and Development Project of Zhejiang Province in China under Grant 2021C03137.

{\small
\bibliographystyle{ieee_fullname}
\bibliography{egbib}
}

\end{document}